\documentclass[11pt, letterpaper]{template}

\usepackage[all]{hypcap}
\usepackage[authoryear, round]{natbib}
\definecolor{citecolor}{HTML}{0071bc}
\usepackage{hyperref}[citecolor=citecolor]
\usepackage{url} 
\hypersetup{
    colorlinks=true,
    citecolor=blue,
    linkcolor=magenta,
    urlcolor=magenta
}

\definecolor{darkblue}{rgb}{0, 0, 0.5}
\usepackage{url}            
\usepackage{booktabs}       
\usepackage{amsfonts}       
\usepackage{nicefrac}       
\usepackage{microtype}      
\usepackage{xcolor}         

\usepackage{lineno}

\usepackage{enumitem}       
\usepackage{subfig}         
\usepackage{multirow}       %
\usepackage{xspace}         
\usepackage{wrapfig}        
\usepackage{makecell}       
\usepackage{colortbl}
\usepackage{bbm}            
\usepackage{adjustbox}

\newboolean{showsection}
\setboolean{showsection}{true}
\makeatletter
\@namedef{ver@everyshi.sty}{}
\makeatother

\definecolor{custom_green}{rgb}{0.0, 0.5, 0.0}
\definecolor{custom_red}{rgb}{1.0, 0.01, 0.24}
\definecolor{custom_blue}{HTML}{C9DAF7}
\definecolor{custom_purple}{HTML}{D9D1E9}
\definecolor{title_blue}{HTML}{204899} 
\definecolor{cite_blue}{HTML}{044dc1}  
\definecolor{cite_purple}{HTML}{7406a7}  
\hypersetup{
    colorlinks = true,
    citecolor = {cite_blue},
    linkcolor = {cite_purple},
    urlcolor = magenta,
}
\usepackage[most,skins,theorems]{tcolorbox}
\tcbset{
  aibox/.style={
    width=\linewidth,
    top=8pt,
    bottom=4pt,
    colback=blue!6!white,
    colframe=black,
    colbacktitle=black,
    enhanced,
    center,
    attach boxed title to top left={yshift=-0.1in,xshift=0.15in},
    boxed title style={boxrule=0pt,colframe=white,},
  }
}
\newtcolorbox{AIbox}[2][]{aibox,title=#2,#1}

\title{Lumina-mGPT 2.0: \\Stand-Alone AutoRegressive Image Modeling}

\author[1,4,6,$\clubsuit$]{Yi Xin}
\author[1,$\clubsuit$]{Juncheng Yan}
\author[1,$\clubsuit$]{Qi Qin}
\author[2,$\clubsuit$]{Zhen Li}
\author[2,$\clubsuit$]{Dongyang Liu}
\author[1]{Shicheng Li}
\author[2]{Victor Shea-Jay Huang}
\author[1]{Yupeng Zhou}
\author[2]{Renrui Zhang}
\author[2]{Le Zhuo}
\author[4]{Tiancheng Han}
\author[4]{Xiaoqing Sun}
\author[3]{Siqi Luo}
\author[5]{Mengmeng Wang}
\author[1]{Bin Fu}
\author[1]{Yuewen Cao}
\author[2]{Hongsheng Li}
\author[3]{Guangtao Zhai}
\author[3]{Xiaohong Liu}
\author[1]{Yu Qiao}
\author[1,$\dag$]{Peng Gao}

\affil[1]{Shanghai AI Laboratory}
\affil[2]{The Chinese University of Hong Kong}
\affil[3]{Shanghai Jiao Tong University}
\affil[4]{Shanghai Innovation Institute}
\affil[5]{Zhejiang University of Technology}
\affil[6]{Nanjing University}

\correspondingauthor{$^\clubsuit$ Equal Contribution.}
\begin{abstract}
\textbf{Abstract} — We present Lumina-mGPT 2.0, a stand-alone, decoder-only autoregressive model that revisits and revitalizes the autoregressive paradigm for high-quality image generation and beyond. 
Unlike existing approaches that rely on pretrained components or hybrid architectures, Lumina-mGPT 2.0 is trained entirely from scratch, enabling unrestricted architectural design and licensing freedom. 
It achieves generation quality on par with state-of-the-art diffusion models such as DALL·E 3 and SANA, while preserving the inherent flexibility and compositionality of autoregressive modeling. 
Our unified tokenization scheme allows the model to seamlessly handle a wide spectrum of tasks—including subject-driven generation, image editing, controllable synthesis, and dense prediction—\textit{within a single generative framework.}
To further boost usability, we incorporate efficient decoding strategies like inference-time scaling and speculative Jacobi sampling to improve quality and speed, respectively.
Extensive evaluations on standard text-to-image benchmarks (e.g., GenEval, DPG) demonstrate that Lumina-mGPT 2.0 not only matches but in some cases surpasses diffusion-based models. 
Moreover, we confirm its multi-task capabilities on the Graph200K benchmark, with the native Lumina-mGPT 2.0 performing exceptionally well.
These results position Lumina-mGPT 2.0 as a strong, flexible foundation model for unified multimodal generation.

\end{abstract}

\begin{document}
\maketitle
\begin{figure*}[!bht]
    \centering
    \begin{picture}(0,295)
    \put(-238,-2){
    \includegraphics[width=1.0\linewidth]{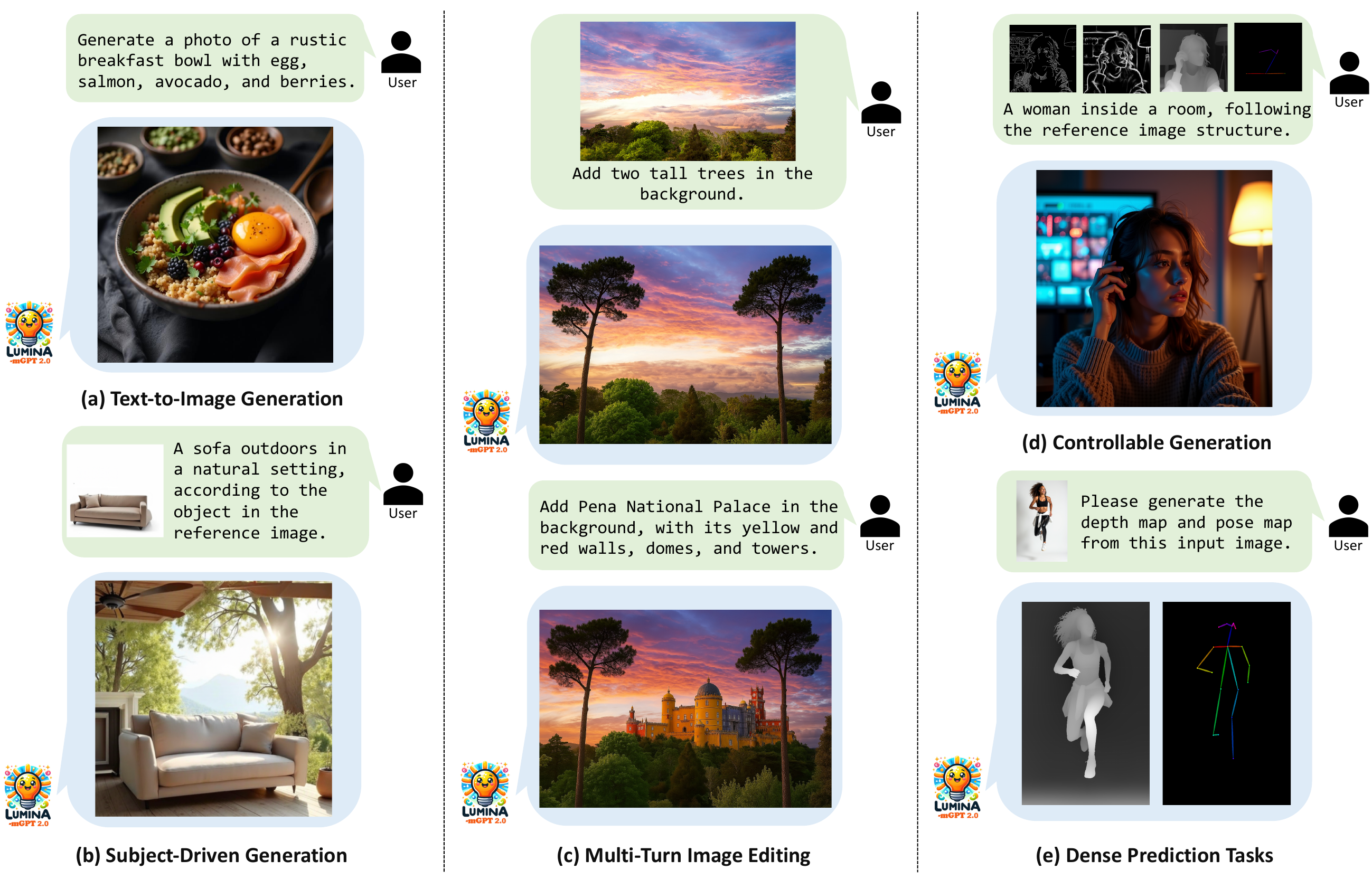}
    }
    \end{picture}
    \vspace{-0.15in}
    \caption{\textbf{Multitask Capabilities of Lumina-mGPT 2.0}. A stand-alone, decoder-only autoregressive model, trained from scratch, that unifies a broad spectrum of image generation tasks.}
    \label{fig:abs_case}
    \vspace{-0.14in}
\end{figure*}
\vspace{-0.5in}
\section{Introduction}
The multimodal autoregressive (AR) paradigm has recently gained widespread attention~\citep{li2024autoregressive,liu2024lumina,sun2024autoregressive,team2024chameleon,liu2024customize,wang2024emu3,han2024infinity}. 
This paradigm, characterized by a unified next-token prediction framework, offers the unique advantage of aligning diverse modalities and tasks within a coherent structure. 
Moreover, its architectural and conceptual similarity to Large Language Models (LLMs) facilitates the seamless application of state-of-the-art training and inference techniques. 
While this paradigm was initially proposed years ago~\citep{pixelrnn, dalle}, it gradually lost prominence as alternative approaches, such as Generative Adversarial Networks (GANs)~\citep{goodfellow2020generative,pan2023drag,karras2019style} and Diffusion Models~\citep{podell2023sdxl,chen2023pixart,LuminaImage2024,zhuo2024lumina,xie2024sana,flux2024}, gained widespread adoption.
However, in 2024, Chameleon~\citep{team2024chameleon} reinvigorated and expanded this paradigm, restoring its promising potential to the forefront of the research community. 
Building upon Chameleon, Lumina-mGPT~\citep{liu2024lumina} emerged as the pioneering open-source initiative to demonstrate that decoder-only AR models can generate high-resolution images with flexible aspect ratios, achieving quality on par with SDXL-level~\citep{podell2023sdxl} diffusion models.

Subsequent developments in this domain have predominantly advanced along two principal directions: modeling adjustments and scaling of resource investments. 
From a modeling perspective, Show-o~\citep{xie2024show} and Transfusion~\citep{zhou2024transfusion} replace traditional autoregressive paradigms in image generation with diffusion-based approaches, creating hybrid diffusion-AR frameworks. 
Concurrently, Janus~\citep{wu2024janus} and Janus-Pro~\citep{chen2025januspro} enhance image understanding by integrating continuous visual encoders (SigLIP~\citep{zhai2023sigmoid}). 
While these modifications bring respective benefits, they also introduce \textit{inconsistencies between visual understanding and generation}, thereby limiting their adaptability for more flexible interactions, such as unified generation. 
In contrast, Emu3~\citep{wang2024emu3} retains a purely AR-based architecture and emphasizes large-scale resource investments. 
Nonetheless, it continues to underperform in terms of image-generation quality and remains underexplored in its potential for unified generation capabilities that are inherently well-suited to pure AR architectures.

In this paper, we introduce \textbf{Lumina-mGPT 2.0}, a pure AR model that achieves performance comparable to state-of-the-art diffusion models such as SANA~\citep{xie2024sana} and DALL·E 3~\citep{betker2023improving} across various text-to-image (T2I) benchmarks. 
Building upon its superior image generation capabilities, we further leverage the inherent strengths of AR architectures to explore unified generation, demonstrating that partial capabilities of GPT-4o image generation (shown in Figure~\ref{fig:abs_case}) can be achieved with modest computational resources (e.g., 64 A100 GPUs over 4 to 5 weeks). 
Notably, in contrast to previous works like Janus-Pro~\citep{chen2025januspro} and Lumina-mGPT~\citep{liu2024lumina}, our approach does not depend on any pretrained weights. 
This not only underscores the efficiency of our method but also empowers us to freely design the model architecture and other components without the constraints of pretrained models or their licensing restrictions.

\section{Related Work}
\label{sec:related_work}
\subsection{Diffusion Models} 
Diffusion models have exhibited remarkable performance in T2I generation. Within this framework, the model is tasked with predicting the Gaussian noise added to continuous latent representations. 
Initially, diffusion models predominantly relied on the U-Net architecture~\citep{rombach2022high,podell2023sdxl}. However, with the growing success of Transformer architectures, the Diffusion Transformer (DiT)~\citep{dit} was introduced as a replacement for the U-Net, marking a significant advancement toward more efficient and scalable diffusion models~\citep{rombach2022high,xie2024sana,zhuo2024lumina,chen2023pixart,betker2023improving,flux2024,liu2025lumina}. 
Building on these foundational models, considerable progress has been made in optimizing sampling quality and efficiency~\citep{song2019generative, song2020denoising, lu2022dpm, yi2024towards}, as well as developing downstream tasks such as controllable generation~\citep{zhang2023adding, zhao2023uni}, subject-driven generation~\citep{ruiz2023dreambooth, xiao2024fastcomposer}, and image editing~\citep{brooks2023instructpix2pix, couairon2023diffedit}.

\subsection{Autoregressive Models} 
Autoregressive (AR) models, leveraging the powerful scaling capabilities of LLMs~\citep{brown2020language,achiam2023gpt,touvron2023llama,liu2024customize,xin2025resurrect}, have been extended to image generation by pioneering works such as VQGAN~\citep{esser2021taming}, CogView~\citep{ding2021cogview}, and Parti~\citep{yu2022scaling}. 
These approaches employ a two-stage process in which an image tokenizer first encodes continuous images into discrete tokens, followed by a Transformer decoder that generates images through next-token prediction. 
Building upon this foundation, subsequent efforts such as LlamaGen~\citep{sun2024autoregressive}, Emu3~\citep{wang2024emu3}, Chameleon~\citep{team2024chameleon}, Lumina-mGPT~\citep{liu2024lumina}, and Loong~\citep{wang2024loong} have scaled up AR models, demonstrating that the generation performance of AR models is comparable to diffusion models. 
Furthermore, recent studies~\citep{li2024autoregressive,fan2024fluid,zhou2024transfusion,xie2024show} have explored hybrid approaches that integrate AR models with diffusion techniques.

\section{Revisiting Lumina-mGPT}
Lumina-mGPT~\citep{liu2024lumina} represents a family of open-source multimodal autoregressive models, which are the first of their kind to be equipped with high-quality, high-resolution, and flexible-aspect-ratio image generation capabilities. 
At the core of Lumina-mGPT are two designs: Flexible-Progressive Supervised FineTuning (FP-SFP), a training recipe with progressively increased image resolutions; and Unambiguous image Representation (Uni-Rep), a dedicated image representation mechanism to handle the inherent 2D-shape ambiguity of 1D-flattened image tokens, and that builds the foundation of Lumina-mGPT's capability to understand and generate images at variable aspect ratios. 
However, while the work has made an important step, it also comes with certain deficiencies:

\begin{figure*}[t]
    \centering
    \includegraphics[width=1.0\linewidth]{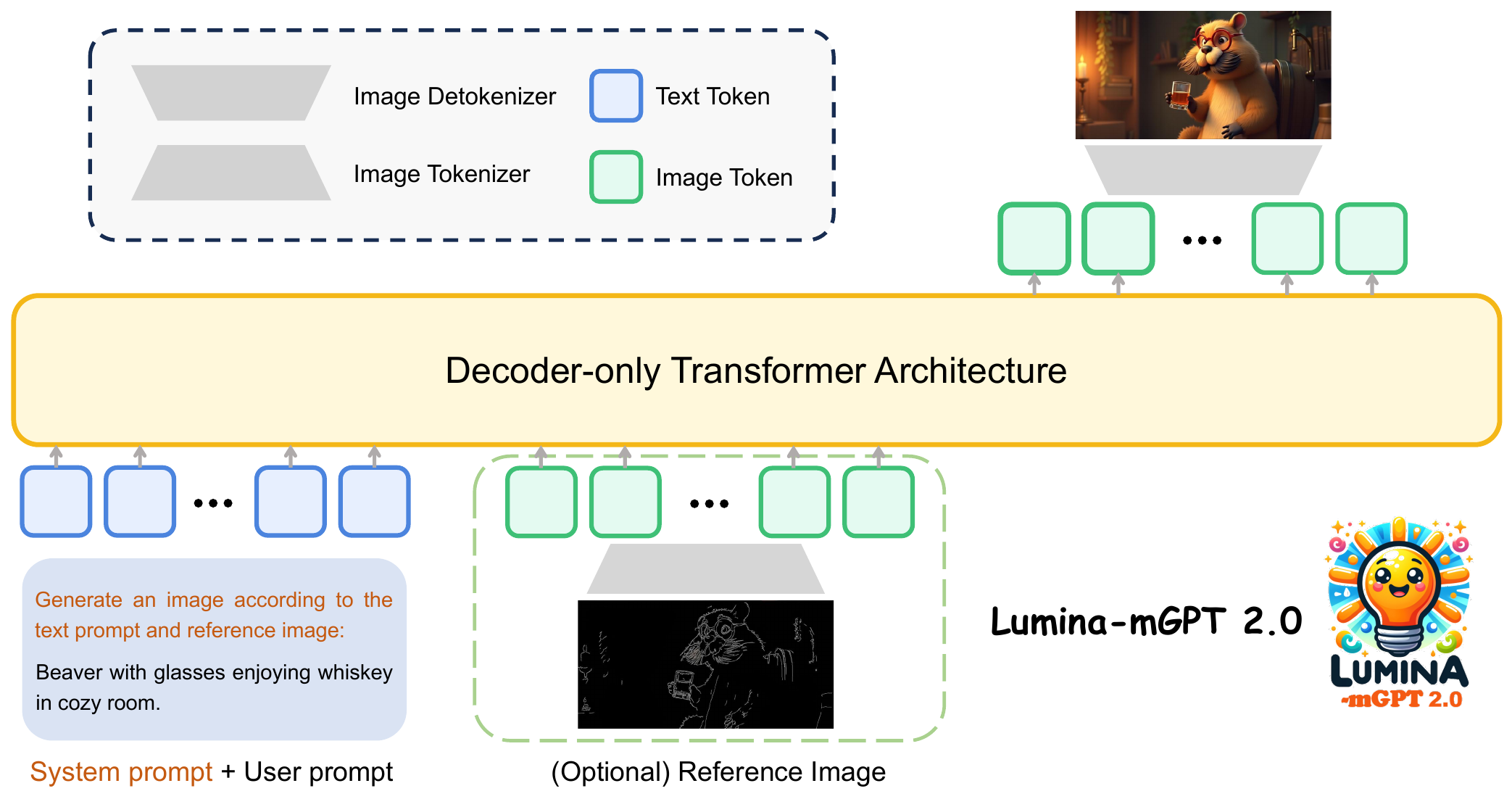}
    \caption{\textbf{Decoder-only Transformer Architecture of Lumina-mGPT 2.0.} This architecture utilizes autoregressive modeling for image synthesis and supports conditional image input, facilitating a wide range of generation tasks through the integration of both text prompts and optional reference images.}
    \label{fig:abs_images}
    \vspace{-0.15in}
\end{figure*}

\vspace{0.1cm} \noindent \textbf{Constraints from Loading Pre-trained Models.} Lumina-mGPT initializes from the pre-trained Chameleon~\citep{team2024chameleon} checkpoint. 
As also faced by other works~\citep{chen2025januspro} that load pre-trained weight, this choice incurs critical limitations: the model architecture and accompanying components (like the image/text tokenizers) must strictly conform to the pre-trained model, which not only impedes customization demands (e.g. creating a different sized model) but also blocks the room for chasing better choices. 
Besides, pre-trained models also impose license restrictions, limiting the usability for broader commercial and real-world deployments.

\vspace{0.1cm} \noindent \textbf{Multitask Conflicts within a Model.} Beyond the fundamental T2I generation task, Lumina-mGPT is required to be separately fine-tuned to extend its capabilities to different downstream generation tasks. 
It treats these tasks as conditional extensions with different checkpoints, rather than integrating them into a unified training paradigm within one framework. 
This separation hinders the effective alignment of multitask objectives with the primary image generation task, limiting the model's overall coherence and efficiency.

\vspace{0.1cm} \noindent \textbf{Limited Inference-time Optimization.} Multi-modal AR models rely on thousands of next-token prediction steps, leading to significant computational overhead and prolonged inference times. 
Notably, numerous inference optimization techniques~\citep{teng2024accelerating,guo2025can} exist, which can substantially reduce sampling time or potentially improve the generation quality. 
However, exploration along this direction was absent in Lumina-mGPT.

\vspace{0.1cm} \noindent \textbf{Inferior Performance Compared to SOTA Models.} Although the image generation capability already achieved a milestone in its field, it still falls behind state-of-art (SOTA) diffusion models, such as Lumina-Image 2.0~\citep{LuminaImage2024}, Sana~\citep{xie2024sana} and DALL·E 3~\citep{betker2023improving}, which leaves room for further improvement.

\section{Lumina-mGPT 2.0}
\label{sec:method}

In this section, we introduce the methodology of our Lumina-mGPT 2.0, which features three characteristics: 1) stand-alone architecuture, 2) unifying diverse generation tasks, and 3) optimized inference strategy.

\subsection{Stand-alone Architecuture}
\vspace{0.1cm} \noindent \textbf{Decoder-only Transformer Trained from Scratch.}
Lumina-mGPT 2.0 retains a similar structural design to its predecessor, Lumina-mGPT \citep{liu2024lumina}, by continuing to utilize the decoder-only Transformer architecture, as shown in Figure~\ref{fig:abs_images}. 
In contrast to Lumina-mGPT, which relies on fine-tuning pre-trained Chameleon 7B and 34B models~\citep{team2024chameleon}, Lumina-mGPT 2.0 is developed as an entirely stand-alone model. 
Specifically, Lumina-mGPT 2.0 
adopts a training-from-scratch paradigm with parameters randomly initialized, leading to several advantages: 1) Bias Reduction: Training from scratch minimizes biases usually inherited from pre-trained models, thereby improving image generation performance. 2) Flexible Architecture: This approach allows for flexible adaptation in model design. 
For example, we offer a more lightweight version of Lumina-mGPT 2.0 with 2B parameters for the T2I community. Additionally, it provides the flexibility to integrate improved image and text tokenizers as needed. 3) Licensing Independence: Without dependence on Chameleon models, Lumina-mGPT 2.0 avoids any potential licensing constraints.  

\begin{figure*}[!t]
    \centering
    \includegraphics[width=1.0\linewidth]{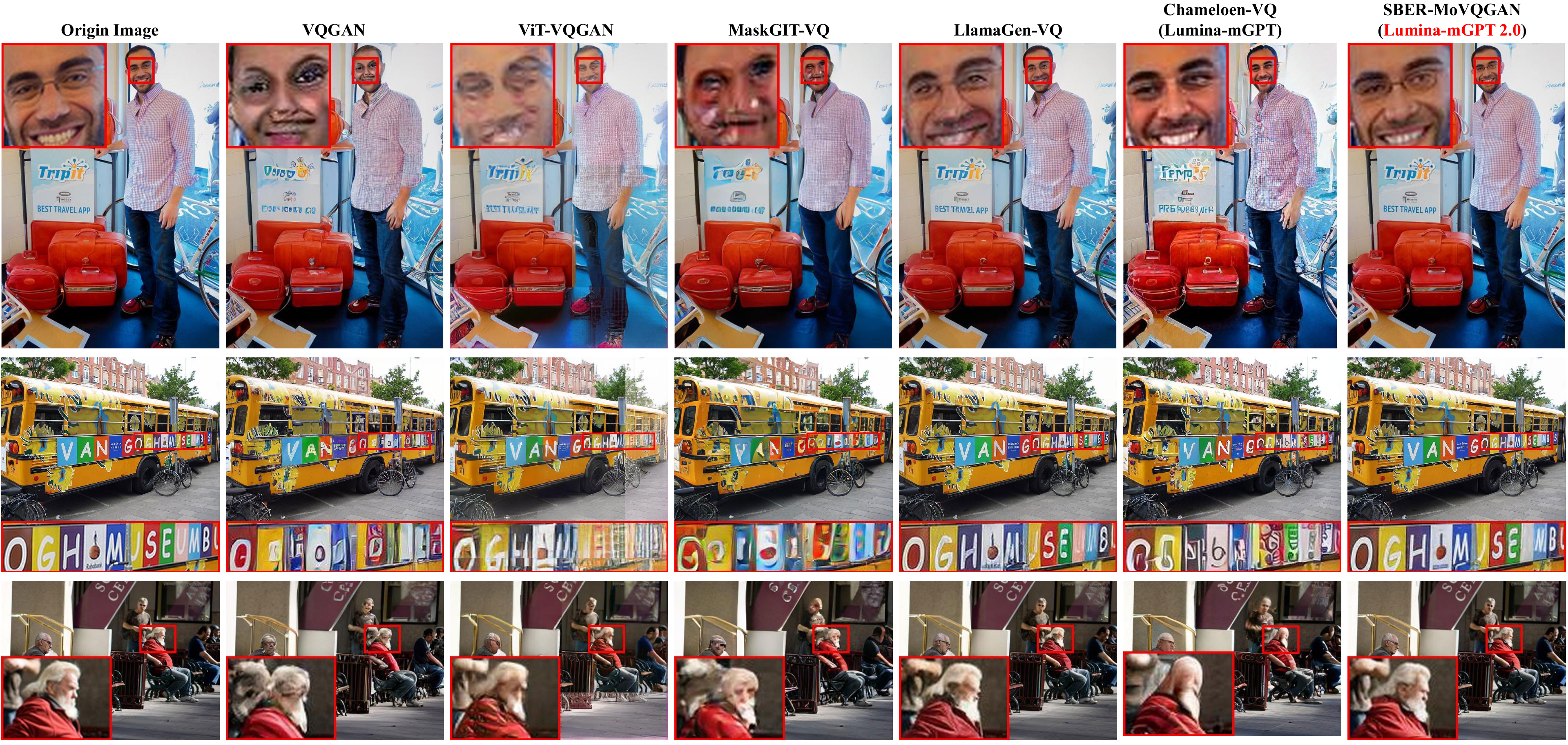}
    \caption{\textbf{Image Reconstruction Results of Different Image Tokenizers.} We present the specific details of selected areas, highlighted with \textcolor{red}{red boxes}, to demonstrate the performance of various image tokenizers.}
\label{fig:vq_img}
\end{figure*}

\begin{table*}[t]
    \centering
    \setlength{\tabcolsep}{10pt}
    \renewcommand\arraystretch{1.18}
    \resizebox{1.0\linewidth}{!}{
        \begin{tabular}{lccccc}
        \toprule
        \textbf{Tokenizer} & \textbf{Year} & \textbf{Ratio} & \textbf{PSNR}$\uparrow$ & \textbf{SSIM}$\uparrow$ & \textbf{LPIPS}$\downarrow$ \\
        \midrule
        VQGAN~\citep{esser2021taming} &2021 &16$\times$16 &18.70 &0.48 &0.17 \\
        ViT-VQGAN~\citep{yuvector} &2022 & 8$\times$8  &18.88 &0.60 &0.16 \\
        MaskGIT-VQ~\citep{chang2022maskgit} &2022 & 16$\times$16 &18.55 &0.47 &0.19\\
        \rowcolor{gray!10} 
        SBER-MoVQGAN~\citep{razzhigaev2023kandinsky} &2023 &8$\times$8 &\textbf{22.77} &\textbf{0.63} &\textbf{0.08}\\
        LlamaGen-VQ~\citep{sun2024autoregressive} &2024  &8$\times$8 &21.91 &0.61 & 0.09\\
        Chameleon-VQ~\citep{team2024chameleon} &2024 &16$\times$16 &18.63 &0.47 &0.18\\  
    \bottomrule
    \end{tabular}
    }
    \caption{\textbf{Reconstruction Performance of Popular Image Tokenizers.} We primarily evaluate the reconstruction metrics (e.g., PSNR, SSIM, and LPIPS) of various image tokenizers on the MS-COCO dataset~\citep{lin2014microsoft}.}
    \label{tab:vqgan}
    \vspace{-0.15in}
\end{table*}

\begin{figure*}[t]
    \centering
    \includegraphics[width=0.93\linewidth]{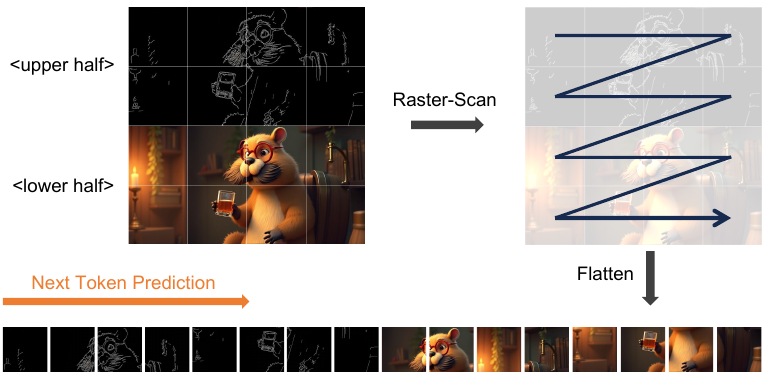}
    \caption{\textbf{Unifying Diverse Generation Tasks with  Autoregressive Raster-Scan Scheme.} The model generates the upper half of the image first (or the reference image given during the inference), which serves as contextual guidance for the generation of the lower half.}
    \label{fig:scan}
    \vspace{0.15cm}
\end{figure*}

\begin{table*}[t]
\centering
\setlength{\tabcolsep}{3.5pt} 
\renewcommand{\arraystretch}{1.8}
\resizebox{1.0\linewidth}{!}{ 
\begin{tabular}{l c c c c c c c}
\toprule
\textbf{Model} & \textbf{Params} &\textbf{Vocabulary Size} & \textbf{Hidden Size} & \textbf{Intermediate Size} & \textbf{Heads} & \textbf{KV Heads} & \textbf{Layers}\\ 
\midrule
Lumina-mGPT 2.0 & 2B &171,385 & 2,048 & 8,192 & 32 & 32 & 32 \\
Lumina-mGPT 2.0 & 7B &171,385 & 4,096 & 11,008 & 32 & 32 & 32 \\
\bottomrule
\end{tabular}
}
\caption{\textbf{Architecture Configuration for Lumina-mGPT 2.0.} The model is scaled from 2B to 7B parameters, with the primary scaling strategy involving an increase in the model's dimension.}
\label{tab:model_hyperparams}
\vspace{-0.1in}
\end{table*}

\begin{table*}[t]
\centering
\renewcommand{\arraystretch}{2.1}  
\begin{tabular}{>{\centering\arraybackslash}m{5cm} m{10.8cm}}  
\hline
\textbf{Text-to-Image Generation}& Generate an image of \{width\}$\times$\{height\} according to the following text prompt: \{\textit{User Text Prompt}\}\\
\hline
\textbf{Subject-Driven Generation} & Generate a dual-panel image of \{width\}$\times$\{height\} where the \textcolor{blue}{$<$upper half$>$} displays: \{\textit{Object Description}\}, while the \textcolor{blue}{$<$lower half$>$} shows the image according to the object and following prompt: \{\textit{Subject Driven Prompt}\}. \\
\hline
\textbf{Image Editing} & Generate a dual-panel image of \{width\}$\times$\{height\} where the \textcolor{blue}{$<$upper half$>$} displays an image: \{\textit{Image Description}\}, while the \textcolor{blue}{$<$lower half$>$} shows an image according to the upper part: \{\textit{Editing Instruction}\} \\ 
\hline
\textbf{Controllable Generation} & Generate a dual-panel image of \{width\}$\times$\{height\} where the \textcolor{blue}{$<$upper half$>$} shows a \textcolor{orange}{$<$Control Task$>$} image, while the \textcolor{blue}{$<$lower half$>$} displays the  image according to the upper part and following prompt: \{\textit{User Text Prompt}\} \\ 
\hline
\makecell[c]{\textbf{Dense Prediction Tasks}} & Generate a dual-panel image of \{width\}$\times$\{height\} where the \textcolor{blue}{$<$upper half$>$} displays the image according to the following description: \{\textit{Image Description}\}, while the \textcolor{blue}{$<$lower half$>$} shows a \textcolor{orange}{$<$Control Task$>$} image according to the upper part. \\
\bottomrule
\end{tabular}
\caption{\textbf{System Prompt for Lumina-mGPT 2.0.} \textcolor{blue}{$<$lower half$>$} and \textcolor{blue}{$<$upper half$>$} are special characters that indicate where the image is placed. \textcolor{orange}{$<$Control Task$>$} represents downstream tasks, including canny edge detection, depth estimation, pose estimation, etc.}
\label{tab:system_prompt}
\end{table*}

\vspace{0.1cm} \noindent \textbf{Resurrect SBER-MoVQGAN Image Tokenizer.} 
The reconstruction quality of the image tokenizer plays a crucial role in determining the upper bound of generation quality. As Lumina-mGPT 2.0 is a stand-alone model, it allows flexible use of tokenizers. 
To achieve high-quality generation, we conduct a comprehensive reconstruction quality analysis of popular image tokenizers used in AR frameworks, including VQGAN~\citep{esser2021taming}, ViT-VQGAN~\citep{yuvector}, MaskGIT-VQ~\citep{chang2022maskgit}, LlamaGen-VQ~\citep{sun2024autoregressive}, SBER-MoVQGAN~\citep{razzhigaev2023kandinsky}, and Chameleon-VQ~\citep{team2024chameleon}, which is employed in Lumina-mGPT. 
The comparative results on the MS-COCO dataset~\citep{lin2014microsoft}, presented in Table~\ref{tab:vqgan} and Figure~\ref{fig:vq_img}, reveal that SBER-MoVQGAN currently stands as the SOTA model for image reconstruction. 
Consequently, Lumina-mGPT 2.0 adopts SBER-MoVQGAN to ensure superior generation performance. 
However, a challenge is the downsampling ratio of 8$\times$8, which leads to a longer image token sequence and increases the inference time and cost.

\vspace{0.1cm} \noindent \textbf{Without Pre-trained Text Encoder.} In Lumina-mGPT 2.0, we represent both text and image data concurrently using a token-based format, as illustrated in Figure~\ref{fig:abs_images}. 
This approach distinguishes Lumina-mGPT 2.0 from some conventional AR methods~\citep{sun2024autoregressive,yu2022scaling}, which typically use a pre-trained text encoder to derive encoded text features and subsequently project these features into the model using an MLP. 
Instead, Lumina-mGPT 2.0 leverages the QwenTokenizer~\citep{wang2024qwen2} to encode text directly into discrete tokens. 
This method simplifies the process, transitioning into a pure next-token prediction paradigm, thereby eliminating the necessity of loading a pre-trained text encoder.

\vspace{0.1cm} \noindent \textbf{Model Scaling.}
To demonstrate the scalability of our stand-alone autoregressive image modeling, we provide two model sizes within the Lumina-mGPT 2.0 family: 2B and 7B. 
The hyperparameters for each model are detailed in Table~\ref{tab:model_hyperparams}. The scaling primarily involves increasing the model's dimension. 
During the experiment, we observe that as model size increases, the convergence speed of training loss accelerates, and the quality of generated images improves substantially in terms of coherence, fine-grained detail, and fidelity to refined prompts, details in Section~\ref{exp:scaling}. These advancements underscore the robust scaling behavior of our model.

\vspace{-0.1cm}
\subsection{Unifying Diverse Generation Tasks}
The autoregressive architecture of Lumina-mGPT 2.0 facilitates the unification of diverse visual tasks within a joint sequence-generation framework. 
Specifically, we leverage a key advantage of the autoregressive approach: its natural sequential ordering of image token generation in a raster scan manner. 
\textbf{\textit{This ensures that the upper portion of an image is generated first, serving as contextual guidance for the subsequent generation of lower regions, as shown in Figure~\ref{fig:scan}}}.

Building upon this property, we incorporate various text-image-to-image tasks into our framework, including subject-driven generation, image editing, controllable generation, and dense prediction. In addition, this special paradigm can also generate image pairs, as shown in Figure~\ref{fig:case}.
Given multiple images, we simply concatenate them vertically into an image grid for joint modeling, thereby ensuring that upper image regions serve as context during generation. 
For controllable generation, the conditioning image is positioned at the top, while the generated output is positioned below. Similarly, for dense prediction tasks like depth estimation, the original image is positioned at the top, and the corresponding label map is placed at the bottom. 
To further distinguish among these task types, we leverage the $<$\textit{system prompt}$>$, as shown in Table~\ref{tab:system_prompt}. 

This unified paradigm facilitates native multi-task training (all tasks treated as text-to-image generation), allowing the model to simultaneously learn various visual tasks without requiring additional architectural changes. 
In accordance with standard T2I training in AR models, the loss is computed exclusively for image tokens, with text tokens remaining unaltered.
\begin{figure*}[t]
    \centering
    \includegraphics[width=0.88\linewidth]{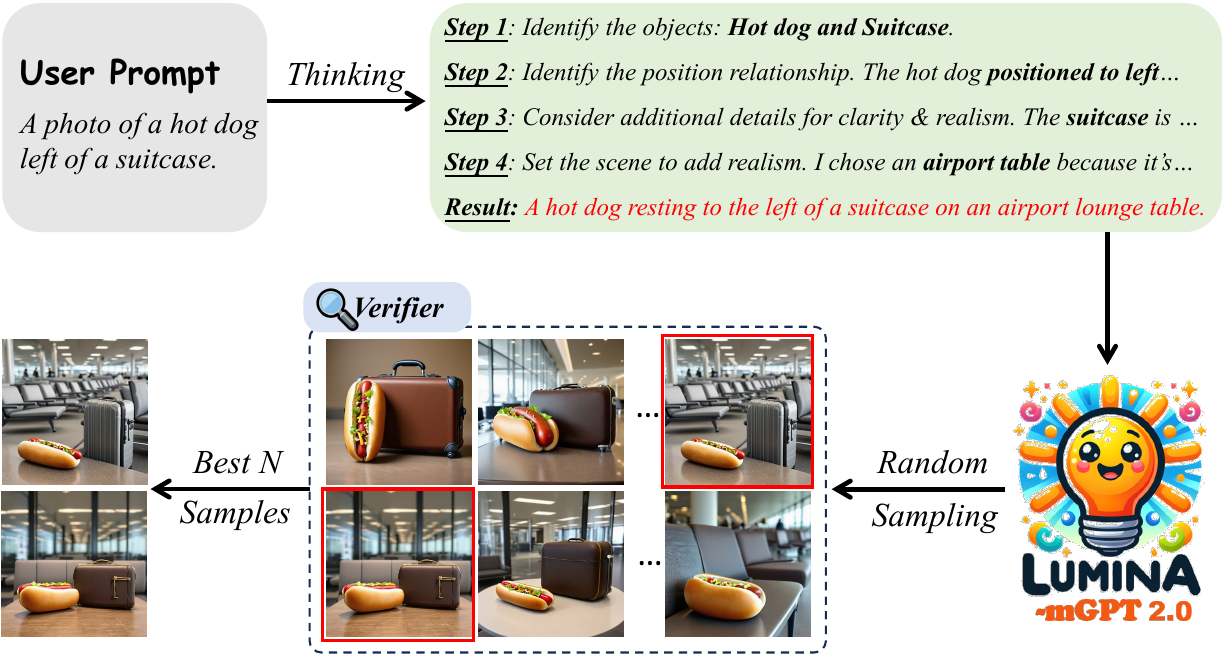}
    \caption{\textbf{Pipeline for High-quality Sampling.} We begin by thinking the user's prompt, elaborating upon it to enhance clarity and coherence. Subsequently, we employ the best-of-N strategy to select the optimal image from the generated candidates.}
\label{fig:inference_thinking_scaling}
\vspace{-0.2cm}
\end{figure*}
At inference time, this formulation offers a flexible and dynamic prompt construction mechanism. 
Users can explicitly specify $<$\textit{system prompt}$>$ to control the task type, determine whether to provide a reference image as the guidance condition, as shown in Figure~\ref{fig:abs_images}. 
This capability seamlessly bridges the gap between controllable and unconditional generation, broadening the native multi-task capabilities of Lumina-mGPT 2.0 as a visual generalist.

\vspace{-0.1cm}
\subsection{Optimized Inference Strategy}
\subsubsection{High-quality Sampling}
\vspace{0.1cm} \noindent \textbf{Thinking Before Generation.} 
Image generation is similar to artistic creation, requiring deep thinking, conceptualization, and iterative refinement before execution. 
However, current image generation models often lack this pre-creative reasoning, instead treating text prompts as direct instructions rather than as an evolving thought process. 
In reality, user prompts are frequently vague, ambiguous, or deficient in crucial details necessary for generating a coherent and meaningful image. 
Motivated by the remarkable progress in Chain-of-Thought (CoT) reasoning within LLMs~\citep{wei2022chain,wang2022self}, we introduce a ``thinking before generation'' paradigm.

\begin{figure*}[!h]
    \centering
    \vspace{-0.2cm}
    \includegraphics[width=0.98\linewidth]{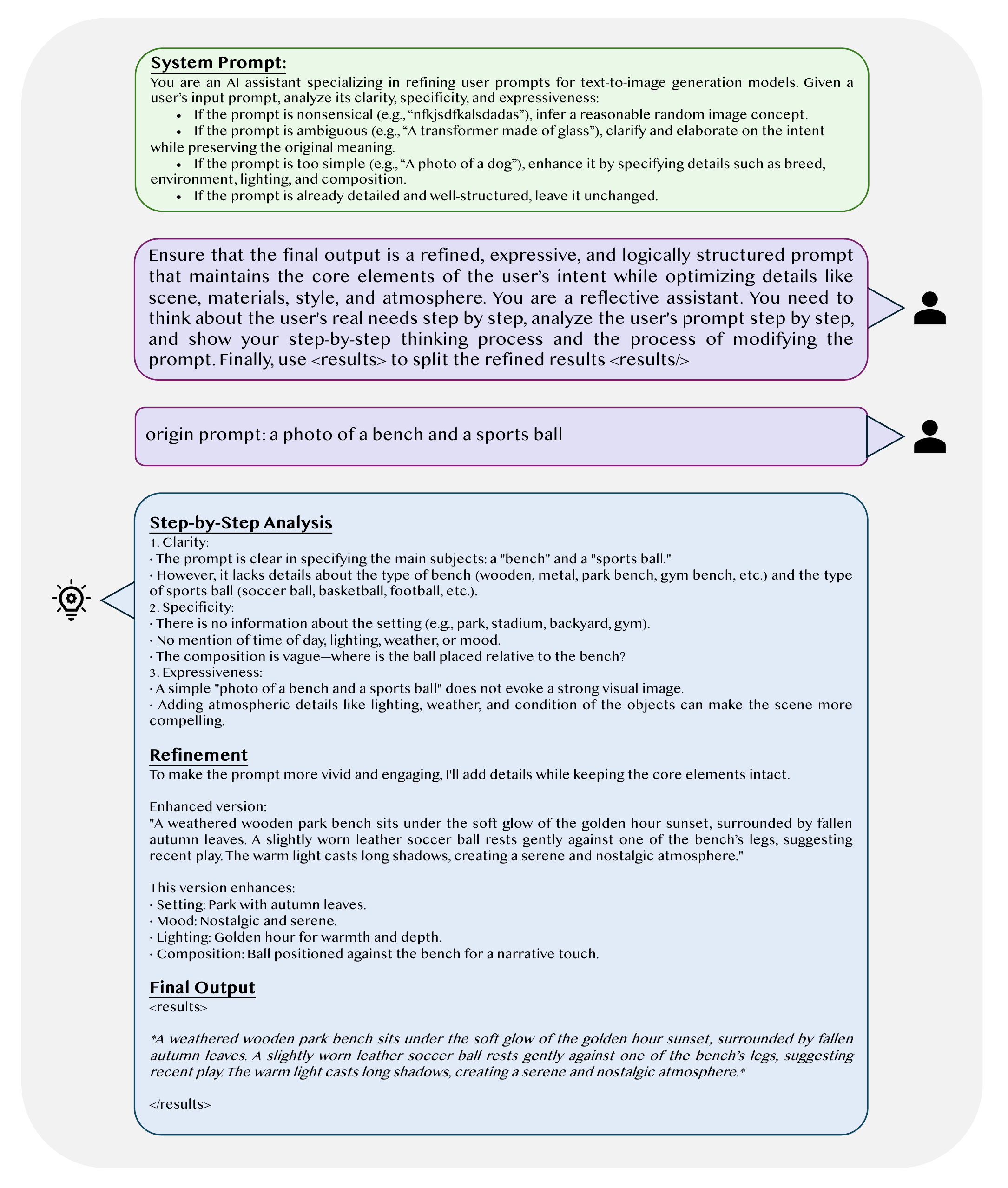}
    \vspace{-0.2cm}
    \caption{\textbf{An Example of Calling GPT-4o to Complete the Thinking Process.} This approach differs from the conventional prompt rewriting process because it centers around a thoughtful, step-by-step thinking process.}
\label{fig:thinking_template}
\vspace{-0.3cm}
\end{figure*}

Specifically, rather than directly feeding the user prompts into Lumina-mGPT 2.0, we first process it through an LLM (GPT-4o). 
The LLM is prompted to systematically analyze and understand the user's underlying intent through step-by-step reasoning, ultimately producing a refined prompt with enhanced coherence, descriptive richness, and clarity. For instance, when encountering a nonsensical prompt, the model infers a plausible interpretation; when faced with ambiguity, it clarifies and elaborates; when the prompt is overly simplistic, it enriches the description by incorporating essential elements, details in Figure~\ref{fig:thinking_template}. 
In this process, the LLM serves as a reasoning engine, progressively refining the prompt in a manner analogous to how an artist iteratively develops their vision. 
By integrating this reflective reasoning process, our approach ensures that the final prompt is not only more structured and expressive but also more faithfully aligned with the user’s original intent.

\vspace{0.1cm} \noindent \textbf{Inference-time Scaling.} Recent studies have begun to investigate inference-time scaling behavior in diffusion models~\citep{ma2025inference,xie2025sana,zhuo2025reflection} as well as hybrid AR and diffusion models~\citep{guo2025can}. 
Building upon these advancements, we take a first step toward exploring pure AR inference-time scaling in our Lumina-mGPT 2.0. Specifically, given a text prompt, Lumina-mGPT 2.0 first generates a diverse set of candidate images in a stochastic manner. 
Subsequently, we employ a best-of-N selection strategy, wherein verifier evaluates the generated images and selects the most optimal one. 
Given the inherent complexity of images and the rich semantic information embedded within text conditions, a more comprehensive assessment of generation quality is essential. 
To this end, we integrate multiple verifiers, including the VQAScore\footnote{VQAScore is a scoring method designed to assess the prompt-following ability of generative models.}~\citep{li2024genai}, LAION-AestheticScore\footnote{LAION-AestheticScore is a scoring method designed for assessing aesthetic quality.}~\citep{schuhmann2022laion}, PickScore\footnote{PickScore is a scoring mechanism aimed at assessing human preferences.}~\citep{kirstain2023pick}.

\subsubsection{Acceleration Sampling}
\vspace{0.1cm} \noindent \textbf{Model Quantization.}
To optimize GPU memory usage and accelerate inference, we apply post-training quantization to the forward decoding module of Lumina-mGPT 2.0 using TorchAo~\citep{torchao}. 
This method quantizes model weights to 4-bit integers with 128-element grouping while maintaining activation tensors in bfloat16 precision to mitigate potential quality degradation. 
Leveraging the native compilation toolkit in PyTorch 2.0~\citep{ansel2024pytorch}, we incorporate quantized operations through torch.compile with reduce-overhead mode, facilitating kernel auto-tuning and static graph optimization. 
Notably, this optimization is achieved without any modifications to the model’s architecture.

\vspace{0.1cm} \noindent \textbf{Speculative Jacobi Decoding.}
We refine the sampling strategy by adopting Speculative Jacobi Decoding (SJD)~\citep{teng2024accelerating}, which integrates deterministic Jacobi\footnote{Jacobi decoding accelerates autoregressive inference without fine-tuning by iteratively predicting multiple tokens until convergence.}~\citep{santilli2023accelerating,song2021accelerating} iterations with stochastic sampling. 
SJD introduces a probabilistic convergence criterion that accepts tokens based on the likelihood ratios between draft and target token distributions, enabling parallel decoding while preserving sample diversity. 

In practice, our goal is to accelerate sampling by jointly leveraging model quantization and SJD. 
However, a key challenge emerges from the inherent necessity of SJD for a dynamic KV cache. 
In traditional autoregressive decoding, the KV cache dynamically grows by appending new key and value tensors as tokens are generated. 
SJD extends this by incorporating iterative refinement, where tokens may be accepted or rejected based on the convergence criterion, requiring a flexible cache to handle variable sequence lengths and token rollbacks. 
This dynamic behavior conflicts with the constraints imposed by compilation operations, such as those in \texttt{torch.compile}, which demand a static, pre-allocated KV cache to generate optimized kernels and minimize recompilation overhead. 

To overcome this, we propose a static KV cache and a static causal attention mask for SJD, enabling compatibility with static compilation frameworks. 
The KV cache pre-allocates fixed-size buffers and uses a pointer-based mechanism to manage the effective sequence length, avoiding dynamic resizing. 
Similarly, a fixed-size attention mask is pre-computed to support the decoding phases, with the mask adjusted during inference using pointers, ensuring efficient and parallel token predictions. 
This design meets the static memory requirements while maintaining the SJD's efficiency.

\begin{table*}[!t]
\centering
\renewcommand{\arraystretch}{1.15}  
\setlength{\tabcolsep}{4.5pt} 
\resizebox{1.0\linewidth}{!}{%
\begin{tabular}{l c cccc cccc}
\toprule
\multirow{2}{*}{\textbf{Methods}} & \multirow{2}{*}{\textbf{\# Params}} & \multicolumn{4}{c}{\textbf{GenEval} $\uparrow$} & \multicolumn{4}{c}{\textbf{DPG} $\uparrow$} \\
\cmidrule(lr){3-6} \cmidrule(lr){7-10}
& & Two Obj. & Counting & Color Attri. & \textbf{Overall} & Entity & Relation & Attribute & \textbf{Overall} \\
\midrule
\multicolumn{10}{l}{\textbf{Diffusion Models}} \\ \hline
SDv1.5~\citep{rombach2022high}      & 0.9B & -  & -  & -  & 0.40  & 74.23  & 73.49  & 75.39  & 63.18 \\ 
Lumina-Next~\citep{zhuo2024lumina} & 1.7B  & 0.49  &  0.38  &  0.15  & 0.46  & 83.78      &89.78     & 82.67      & 75.66  \\ 
SDv2.1~\citep{rombach2022high}     & 0.9B & 0.51  & 0.44  & 0.50  & 0.47  & -  & -  & -  & 68.09  \\ 
PixArt-$\alpha$~\citep{chen2023pixart} & 0.6B & 0.50 & 0.44  & 0.07  & 0.48  & 79.32  & 82.57  & 78.60  & 71.11\\ 
SDXL~\citep{podell2023sdxl}  & 2.6B & 0.74  & 0.39  & 0.23  & 0.55  & 82.43  & 86.76  & 80.91  & 74.65     \\ 
SD3-medium~\citep{esser2024scaling} & 2B   & 0.74  & 0.63  & 0.36     & 0.62  &  91.01      &   80.70      & 88.83      & 84.08 \\ 
Sana-1.6B~\citep{xie2024sana} & 1.6B  & 0.77  &  0.62  &  0.47  & 0.66  & \textcolor{blue}{91.50}     &\textcolor{blue}{91.90}     & \textcolor{blue}{88.90}      & \textcolor{blue}{84.80} \\ 
DALL-E3~\citep{betker2023improving}   & -  & \textcolor{red}{0.87}    & 0.47     & 0.45  & 0.67 &89.61 & 90.58 &88.39 & 83.50 \\ 
OmniGen~\citep{xiao2024omnigen}  & 3.8B  & \textcolor{blue}{0.86} & \textcolor{blue}{0.64} & \textcolor{blue}{0.55} & \textcolor{blue}{0.70} &  - & - & - & - \\ 
Lumina-Image 2.0~\citep{LuminaImage2024} & 2.6B  & \textcolor{red}{0.87} & \textcolor{red}{0.67} & \textcolor{red}{0.62} &\textcolor{red}{0.73} &\textcolor{red}{91.97} &\textcolor{red}{94.85} &\textcolor{red}{90.20} &\textcolor{red}{87.20}\\ 
\midrule
\multicolumn{10}{l}{\textbf{AutoRegressive Models}} \\ \hline
LlamaGen~\citep{sun2024autoregressive}    & 0.8B & 0.34  & 0.21  & 0.04  & 0.32  & -      & -      & -      & 65.16  \\ 
Chameleon~\citep{team2024chameleon}   & 7B   & -     & -     & -     & 0.39  & -      & -      & -      & - \\ 
Show-o~\citep{xie2024show}     & 1.3B & 0.52  & 0.49  & 0.28  & 0.53  & -      & -      & -      & 67.48     \\ 
Emu3~\citep{wang2024emu3}        & 8.0B & 0.71 & 0.34 & 0.21 & 0.54 & 86.68 & 90.22 & 86.84 & 80.60 \\ 
Infinity~\citep{han2024infinity}  & 2B  & 0.85 & - & 0.57 & 0.73 & - & 90.76 & - & 83.46\\ 
Janus-Pro-1B~\citep{chen2025januspro}  & 1.5B  & 0.82 & 0.51 & 0.56 &  0.73 & 88.63 & 88.98 & 88.17 &  82.63 \\ 
Janus-Pro-7B~\citep{chen2025januspro} & 7B  & \textcolor{blue}{0.89} & \textcolor{red}{0.59} & \textcolor{blue}{0.66} &  \textcolor{red}{0.80} & \textcolor{blue}{88.90} & 89.32 & \textcolor{red}{89.40} &  \textcolor{blue}{84.19}\\ 
Lumina-mGPT~\citep{liu2024lumina} & 7B &0.77 &0.27 &0.32 &0.56 &86.60 &\textcolor{blue}{91.29} &84.61 &79.70\\
\rowcolor{green!10} 
\textbf{Lumina-mGPT 2.0}$^\dagger$ & \textbf{2B} &0.83 &0.50 &0.54 &0.68 &87.37 &90.03 &84.79 &82.05\\
\rowcolor{green!10} 
\textbf{Lumina-mGPT 2.0}$^\dagger$ & \textbf{7B} &\textcolor{red}{0.92} &\textcolor{blue}{0.57} &\textcolor{red}{0.72} &\textcolor{red}{0.80} &\textcolor{red}{88.94} &\textcolor{red}{91.70} &\textcolor{blue}{88.08} &\textcolor{red}{84.30}\\

\bottomrule
\end{tabular}%
}
\caption{\textbf{Performance Comparison Across Different T2I Models on GenEval~\citep{ghosh2024geneval} and DPG~\citep{hu2024ella} Benchmark.} $\dagger$ indicates that GenEval is the result of our sampling optimization strategy. We highlight the \textcolor{red}{best} and the \textcolor{blue}{second} results for diffusion and AR separately.}
\label{tab:performance_comparison}
\end{table*}

\begin{figure*}[!t]
    \centering
    \vspace{-0.4cm}
    \includegraphics[width=0.96\linewidth]{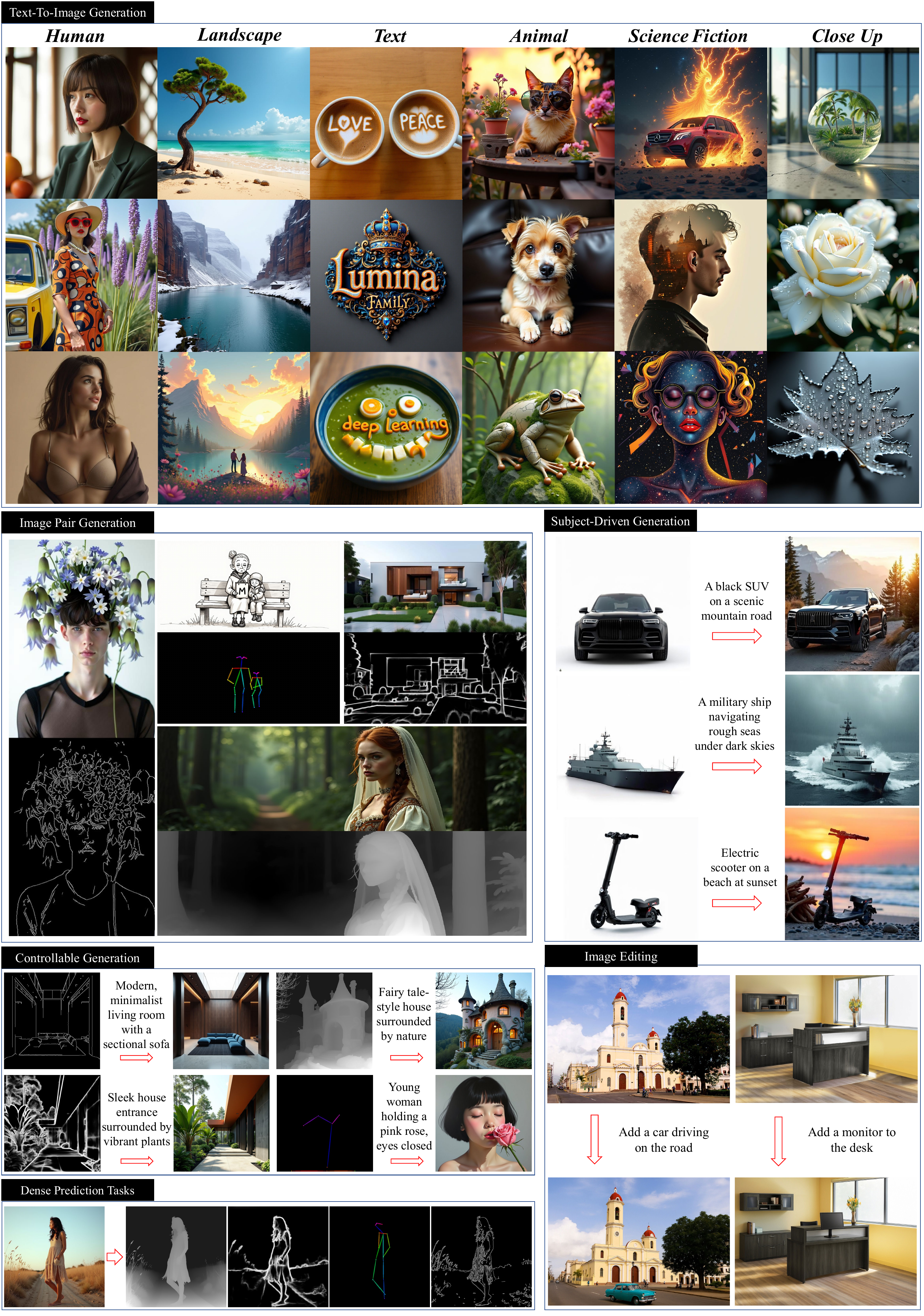}
    \caption{\centering \textbf{Text-to-image Generation and Multi-task Generation Results of Lumina-mGPT 2.0.}}
    \label{fig:case}
    \vspace{-0.1cm}
\end{figure*}

\section{Experiment}
\label{sec:exp}
\subsection{Implement Details}
\vspace{0.1cm} \noindent \textbf{Training Data.} 
Our T2I training dataset is a subset extracted from Lumina-Image 2.0~\citep{LuminaImage2024}, consisting of both real and synthetic data. This dataset has undergone meticulous filtering and re-captioning using OmniCaptioner~\citep{lu2025omnicaptioner}. 
For the multi-task training, different datasets are employed: the subject-driven generation task utilizes Subject200K~\citep{tan2024omini}, the editing task makes use of OminiEdit~\citep{wei2024omniedit}, and the controllable generation and dense prediction tasks are developed by randomly sampling 200K entries from the T2I dataset.

\vspace{0.1cm} \noindent \textbf{Training Details.} Recent research~\citep{LuminaImage2024} has demonstrated that employing a hierarchical data significantly enhances performance, particularly in high-fidelity image synthesis. 
In response, our training pipeline follows a hierarchical three-phase strategy: we first pre-train the model at 256px with a learning rate of 2e-4, followed by fine-tuning at 512px and 768px with the learning rate reduced to 2e-5. The global batch size is dynamically adjusted between 512 and 1,024 through gradient accumulation. 
The training data for each phase of Lumina-mGPT 2.0 is structured as follows: 50M data for the 256px stage, 19M data for the 512px stage, and 8M data for the 768px stage. As the dataset size decreases, the quality of the data progressively improves.  All training is conducted using distributed training across 64 A100 GPUs.

\vspace{0.1cm} \noindent \textbf{Evaluation Details.} We employ a variety of T2I evaluation benchmark to assess Lumina-mGPT 2.0, including GenEval~\citep{ghosh2024geneval}, and DPG~\citep{hu2024ella}, with test sets containing 553 and 1,065 prompts, respectively. These metrics primarily focus on evaluating text-image alignment. Furthermore, we perform comprehensive ablation studies using GenEval to validate the effectiveness of our inference strategy. For multi-task evaluation, we incorporate the benchmark proposed by VisualCloze~\citep{li2025visualcloze}, which enables us to evaluate the capabilities of controllable generation and subject-driven generation using 1,000 samples, respectively.

\newcommand{\best}[1]{\textbf{#1}}
\begin{table*}
  [!t] \small
  \centering
  \renewcommand{\arraystretch}{1.15}
  \setlength{\tabcolsep}{2mm}    
  \resizebox{1.0\linewidth}{!}{
  \begin{tabular}{clccccccc}
    \toprule
    \multirow{2}{*}{\textbf{Condition}}               & \multirow{2}{*}{\textbf{Method}}  &  \multicolumn{2}{c}{\textbf{Controllability}}                          & \multicolumn{4}{c}{\textbf{Quality}} & \textbf{Text Consistency}                     \\
    \cmidrule(lr){3-4} \cmidrule(lr){5-8} \cmidrule(lr){9-9} 
     &  & $\text{F1} \uparrow$ & $\text{RMSE} \downarrow$ & $\text{FID}\downarrow$               & $\text{SSIM}\uparrow$               & $\text{MAN-IQA}\uparrow$            & $\text{MUSIQ}\uparrow$               & $\text{CLIP-Score}\uparrow$          \\
     \midrule \multirow{6}{*}{Canny} & \color{gray} ControlNet~\citep{zhang2023adding} & \color{gray} {0.13} & \color{gray} - & \color{gray} 46.06  & \color{gray} 0.34 & \color{gray} 0.31 & \color{gray} 45.45 & \color{gray} 34.10 \\
     & \color{gray} OminiControl~\citep{tan2024omini}  & \color{gray} 0.47 & \color{gray} - & \color{gray} 29.58 & \color{gray} 0.61 & \color{gray} 0.44 & \color{gray} 61.40 & \color{gray} 34.40 \\
     & OneDiffusion~\citep{le2024diffusiongenerate} & 0.39 & - & \textcolor{blue}{32.76} & \textcolor{red}{0.55} & 0.46 & 59.99 & \textcolor{red}{34.99} \\
     & OmniGen~\citep{xiao2024omnigen} & \textcolor{blue}{0.43} & - &  51.58  &  {0.47} & \textcolor{blue}{0.47} & 62.66 & 33.66  \\
     & Lumina-mGPT~\citep{liu2024lumina} & 0.16 & - & 85.03 & 0.23 & \textcolor{red}{0.48} & \textcolor{red}{70.78} & 28.18\\
     & \cellcolor{green!10}\textbf{Lumina-mGPT 2.0} & \cellcolor{green!10}\textcolor{red}{0.49} & \cellcolor{green!10}- & \cellcolor{green!10}\textcolor{red}{30.89} & \cellcolor{green!10}\textcolor{blue}{0.54} & \cellcolor{green!10}0.42 & \cellcolor{green!10} \textcolor{blue}{63.18} & \cellcolor{green!10}\textcolor{blue}{34.44} \\
    \midrule \multirow{6}{*}{Depth}          & \color{gray} ControlNet~\citep{zhang2023adding} & \color{gray} - & \color{gray} 23.70 & \color{gray} 36.83 & \color{gray} 0.41 & \color{gray} 0.44 & \color{gray} 60.17 & \color{gray} 34.49 \\
     & \color{gray} OminiControl~\citep{tan2024omini}& \color{gray} - & \color{gray} 21.44 & \color{gray} 36.23 & \color{gray} 0.52 & \color{gray} 0.44 & \color{gray} 60.18 & \color{gray} 34.08 \\
     & OneDiffusion~\citep{le2024diffusiongenerate} & - & \textcolor{red}{10.35} & \textcolor{blue}{39.03} & \textcolor{red}{0.49} & \textcolor{red}{0.49} & 60.49 & \textcolor{red}{34.71}   \\
     & OmniGen~\citep{xiao2024omnigen} & - &    \textcolor{blue}{15.07}    &  86.08   &    0.26  & \textcolor{red}{0.49} & \textcolor{blue}{64.90} & 29.72   \\
     & Lumina-mGPT~\citep{liu2024lumina} & - & 15.71  & 61.44 & \textcolor{blue}{0.34} & 0.38 & \textcolor{red}{69.72} & 31.58 \\
     & \cellcolor{green!10}\textbf{Lumina-mGPT 2.0} & \cellcolor{green!10}- & \cellcolor{green!10}17.42 & \cellcolor{green!10}\textcolor{red}{36.52} & \cellcolor{green!10}\textcolor{red}{0.49} & \cellcolor{green!10}\textcolor{blue}{0.39} & \cellcolor{green!10}59.52 & \cellcolor{green!10}\textcolor{blue}{34.03}  \\
    \bottomrule
  \end{tabular}
  }
  \caption{\textbf{Quantitative Comparison on Controllable Generation.} 
  The methods that train a specialist for each task are marked as {\color{gray}{gray color}}. 
  Except for these methods, we highlight the \textcolor{red}{best} and the \textcolor{blue}{second} results.}
  \label{tab:control}
\end{table*}
\subsection{Comparison with State-of-the-arts}
\subsubsection{Quantitative Performance}
\label{sec:t2i_bench}
\vspace{0.1cm} \noindent \textbf{Text-to-image Generation.} We compare Lumina-mGPT 2.0 with the advanced T2I generation methods across two benchmarks in Table~\ref{tab:performance_comparison}. Our model achieves performance comparable to or exceeding that of both AR-based and diffusion-based models, including Emu3~\citep{wang2024emu3}, Janus Pro~\citep{chen2025januspro}, and even Lumina-Image 2.0~\citep{LuminaImage2024}. Notably, Lumina-mGPT 2.0 achieves a GenEval score of 0.80, positioning it among the top-tier generative models available. It particularly excels in GenEval tests related to ``Two Objects'' and ``Color Attribute''. Additionally, Lumina-mGPT 2.0 attains a DPG score of 84.30, surpassing the previously established upper limits for AR-based models.

\begin{wraptable}{r}{8.5cm}
    \centering
    \vspace{-0.14in}
    \renewcommand{\arraystretch}{1.15}
    \setlength{\tabcolsep}{2.3mm}
    \resizebox{1.0\linewidth}{!}{
    \begin{tabular}{lccc}
        \toprule
        {\textbf{Method}} & {\textbf{DINOv2 $\uparrow$}} & {\textbf{CLIP-I $\uparrow$}} & {\textbf{CLIP-T $\uparrow$}} \\
        \midrule
        \color{gray} OminiControl~\citep{tan2024omini} & \color{gray} 73.17 & \color{gray} 87.70 & \color{gray} 33.53  \\
        OneDiffusion~\citep{le2024diffusiongenerate} & \textcolor{blue}{73.88} & \textcolor{blue}{86.91} & \textcolor{red}{34.85} \\
        OmniGen~\citep{xiao2024omnigen} & 67.73 & 83.43 & \textcolor{blue}{34.53}  \\
        Lumina-mGPT~\citep{liu2024lumina} & 60.94 & 70.63 & 30.16 \\
        \midrule
        \cellcolor{green!10}\textbf{Lumina-mGPT 2.0} & \cellcolor{green!10}\textcolor{red}{76.60} & \cellcolor{green!10}\textcolor{red}{87.37} & \cellcolor{green!10} 33.90  \\        
        \bottomrule
    \end{tabular}
    }
    \caption{\textbf{Quantitative Comparison for Subject-Driven Image Generation.} We report clip scores on text alignment and style consistency. 
    Specialists are shaded in {\color{gray}{gray}}.
    Among the remaining methods, we highlight the \textcolor{red}{best} and the \textcolor{blue}{second} results.}
    \label{tab:subject}
\end{wraptable}
\vspace{0.1cm} \noindent \textbf{Native Multi-task.} We mainly evaluate the capabilities of Lumina-mGPT 2.0 in controllable generation, as detailed in Table~\ref{tab:control}, and subject-driven generation, as detailed in Table~\ref{tab:subject}. The results reveal that Lumina-mGPT 2.0 excels as a versatile generalist model. 
In the realm of controllable generation, it achieves top-tier performance, exhibiting high structural adherence in both Canny and Depth conditions while maintaining excellent image quality and text consistency.
In subject-driven tasks, Lumina-mGPT 2.0 surpasses all competitors in maintaining subject identity and achieves impressive results in image consistency and text alignment.

\begin{figure*}[!t]
    \centering
    \vspace{-0.4cm}
    \includegraphics[width=0.955\linewidth]{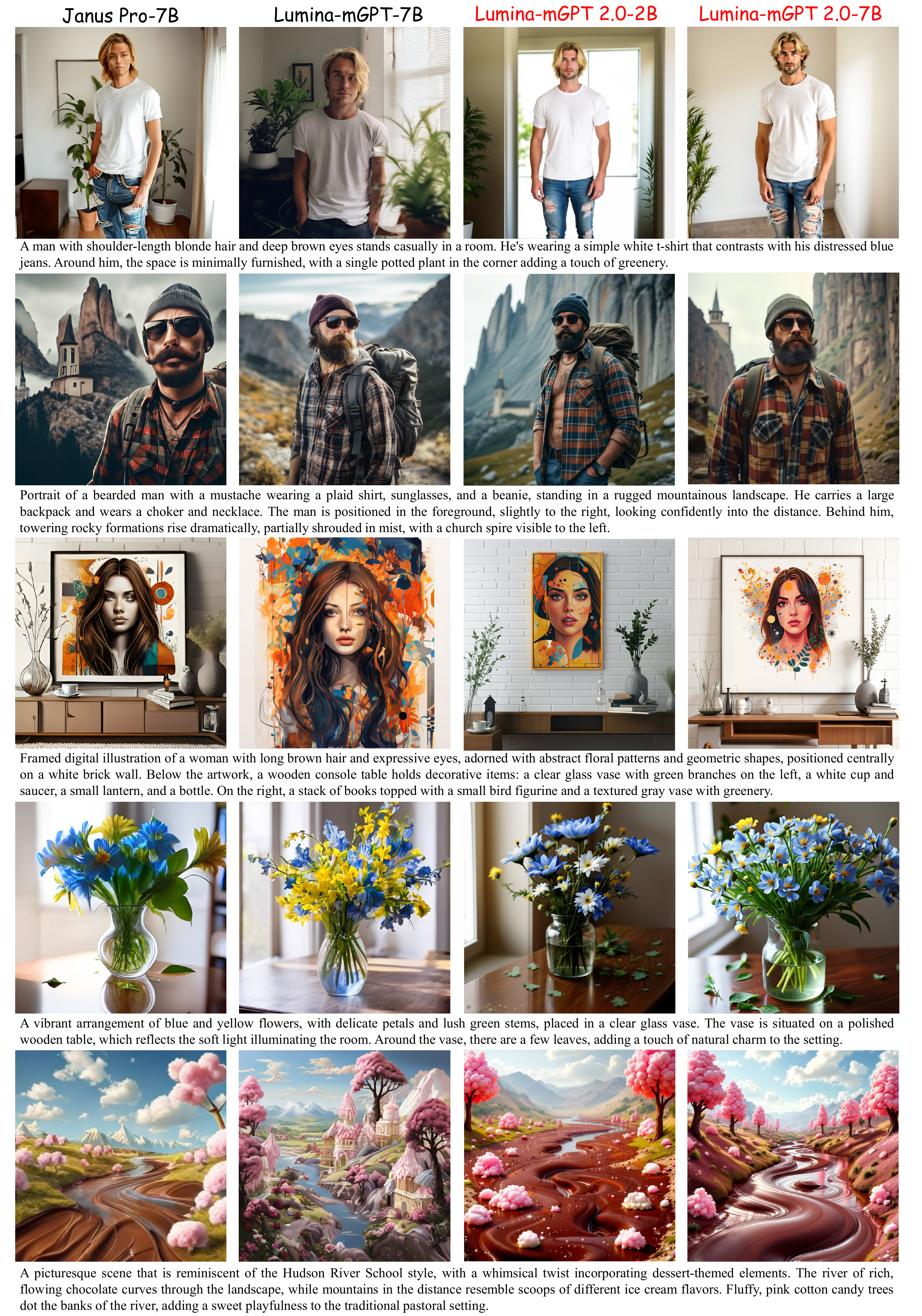}
    \caption{\textbf{Visual Comparison of Text-to-Image Among Lumina-mGPT 2.0, Lumina-mGPT, and Janus Pro.}}
\label{fig:compare}
\vspace{-0.1cm}
\end{figure*}

\begin{figure*}[!t]
    \centering
    \includegraphics[width=1.0\linewidth]{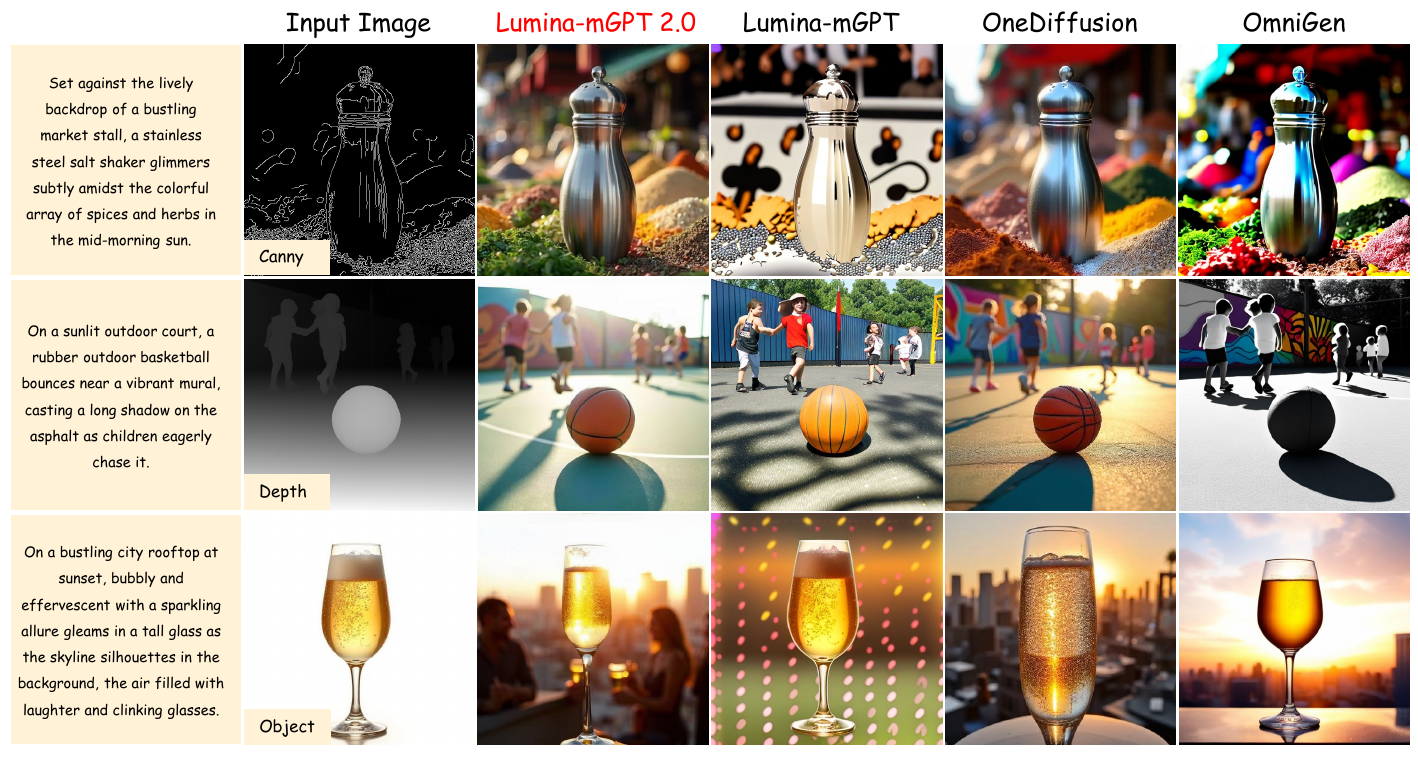}
    \caption{\textbf{Visual Comparison of Controllable/Subject-Driven Generation Among Lumina-mGPT 2.0, Lumina-mGPT, OneDiffusion, and OmniGen.} The control inputs includes Canny (first row) and Depth (second row).}
\label{fig:compare_2}
\end{figure*}

\subsubsection{Qualitative Performance}
\vspace{0.1cm} \noindent \textbf{Text-to-image Generation.} 
We present the T2I generation results in Figure~\ref{fig:case}, demonstrating the model’s proficiency in synthesizing high-fidelity visual content across a broad spectrum of categories. Lumina-mGPT 2.0 effectively generates photorealistic human, breathtaking landscapes, and intricate text-based designs with exceptional detail. Furthermore, it excels in rendering lifelike animals, imaginative science fiction scenes, and highly detailed close-up shots. These results highlight the model’s capability to accurately interpret prompts, capturing rich textures, dynamic lighting effects, and compelling compositions.

In addition, we conduct a comparative analysis of T2I results from Lumina-mGPT 2.0 against those produced by Janus Pro and its predecessor, Lumina-mGPT, as illustrated in Figure~\ref{fig:compare}. 
Lumina-mGPT 2.0 demonstrates significant improvements in realism, detail, and coherence over both its predecessor and Janus Pro. 
The generated images exhibit sharper textures, more precise lighting, and enhanced composition, making them visually compelling and more aligned with natural aesthetics. 
Interestingly, these findings differs from the conclusions drawn from Table~\ref{tab:performance_comparison}, where Lumina-mGPT 2.0 and Janus Pro demonstrate comparable performance on the GenEval benchmark. 
This benchmark primarily relies on VLM models to assess the alignment between text and images, without explicitly considering the quality and aesthetics of image generation.

\vspace{0.1cm} \noindent \textbf{Native Multi-task.} Beyond T2I generation, Lumina-mGPT 2.0 demonstrates remarkable multi-task capabilities, as illustrated in Figure~\ref{fig:case}.
The results demonstrate that Lumina-mGPT 2.0 inherently supports a wide range of image-to-image generation tasks, including subject-driven generation, image editing and controllable generation (e.g., canny-to-image, depth-to-image, pose-to-image, hed-to-image), all without requiring additional modules~\citep{li2024controlar,yao2024car} or additional fine-tuning stage~\citep{chung2025fine}. Furthermore, Lumina-mGPT 2.0 efficiently generates task-specific image pairs to augment training datasets for image-to-image task training of other models, while simultaneously providing robust support for various dense prediction tasks.

In addition, we conduct multi-task generative visual comparisons with other models, including Lumina-mGPT, OneDiffusion, and OmniGen, as illustrated in Figure~\ref{fig:compare_2}. 
Lumina-mGPT 2.0 showcases impressive performance in both controllable generation and subject-driven generation tasks.

\subsection{Ablation Study}
\begin{figure}[!t]
    \centering
    \includegraphics[width=0.7\linewidth]{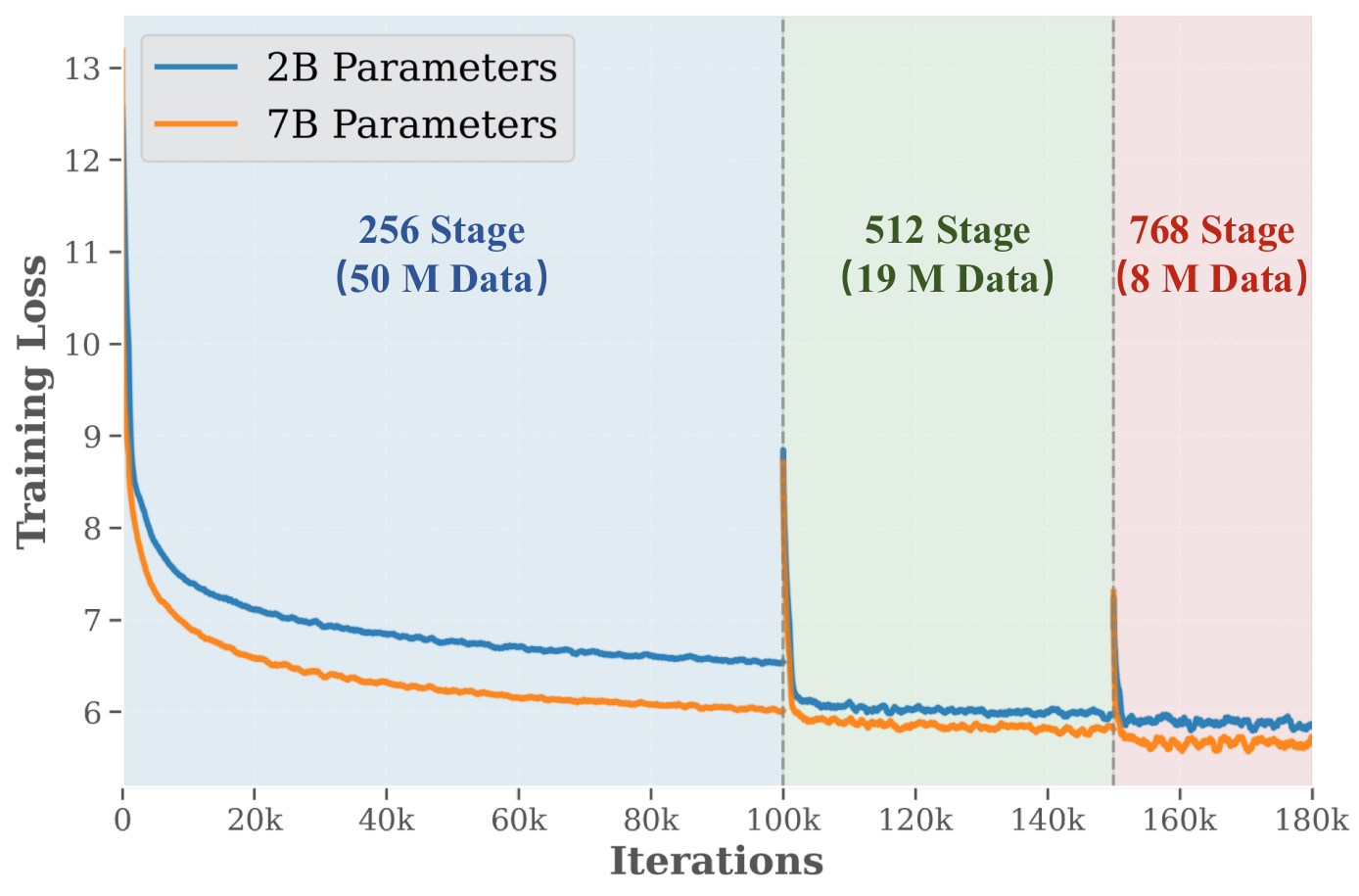}
    \caption{\textbf{Three-stage Training Loss Curves for Lumina-mGPT 2.0.} The training loss curves compare the performance of two model sizes, 2B and 7B parameters. The 7B model consistently achieves lower training loss than the 2B model throughout training, indicating better convergence.}
    \label{fig:scaling_loss}
\end{figure}

\vspace{0.1cm} \noindent \textbf{Effect of Model Growth.} 
\label{exp:scaling}
In Lumina-mGPT 2.0, we scale the model from 2B to 7B parameters. To assess the impact of this scaling, we conduct an analysis from three perspectives: (1) Benchmark Performance: As shown in Table~\ref{tab:performance_comparison}, the 7B model outperforms the 2B variant across multiple benchmarks, including GenEval and DPG. (2) Training Loss: The larger 7B model exhibits a significantly faster training loss convergence compared to smaller 2B model, as shown in Figure~\ref{fig:scaling_loss}. (3) Visual Quality: The generated results of the 7B model demonstrate greater stability, enhanced visual fidelity and richer details, as shown in Figure~\ref{fig:compare}.

\vspace{0.1cm} \noindent \textbf{Effect of Thinking Before Generation.} Before image generation, we call the GPT-4o API to conduct an in-depth analysis of input prompts, fully understand their meaning, and produce enhanced prompts. Examples of the step-by-step thinking process and the resulting enhanced prompts are provided in Figure~\ref{fig:thinking_template}. To evaluate the effectiveness of this approach, we conduct an ablation experiment on GenEval benchmark, as shown in Table~\ref{tab:geneval}. The results indicate that the prompts after the thinking process yield an average improvement of 4\%, with a notable enhancement in position and color attribution ability, each showing a 8\% improvement. These findings suggest that this method more effectively supports the image generation process, aligning more closely with the user’s intentions.

\renewcommand\arraystretch{1.25}
\begin{table*}[!t]
    \centering
    \scalebox{0.734}{
    \begin{tabular}{l|c|cccccc|c}
    \toprule
    \textbf{Method} & \textbf{\# Params.} & \textbf{Single Obj.} & \textbf{Two Obj.} & \textbf{Counting} & \textbf{Colors} & \textbf{Position} & \textbf{Color Attri.} & \textbf{Overall}\\
    \midrule
    LlamaGen~\citep{sun2024autoregressive}             & 0.8B  &0.71 &0.34 &0.21 &0.58 &0.07 &0.04 & 0.32\\
    Emu-3~\citep{wang2024emu3}                         & 8B  &0.98 &0.71 &0.34 &0.81 &0.17 &0.21 &0.54  \\
    Janus-Pro-7B~\citep{chen2025januspro}              & 7B  &\textcolor{blue}{0.99} &\textcolor{blue}{0.89} &\textcolor{red}{0.59} &\textcolor{red}{0.90} &\textcolor{red}{0.79} &\textcolor{blue}{0.66} &\textcolor{red}{0.80}  \\
    Lumina-mGPT~\citep{liu2024lumina}                  & 7B  &0.98 &0.77 &0.27 &0.82 &0.17 &0.32 &0.56   \\
    \midrule
    \textbf{Lumina-mGPT-2.0} &7B &\textcolor{blue}{0.99} &0.87 &0.44 &0.85 &0.44 &0.54 &0.69\\
    \rowcolor{green!10} 
    \textbf{\quad + Thinking Before Generation} & 7B &\textcolor{red}{1.00} &0.87 &0.49 &0.85 &0.52 &0.62 &0.73\\
    \rowcolor{green!10} 
    \textbf{\quad \quad + Inference-Time Scaling}    & 7B  &\textcolor{red}{1.00} &\textcolor{red}{0.92} &\textcolor{blue}{0.57} &\textcolor{blue}{0.88} &\textcolor{blue}{0.70} &\textcolor{red}{0.72} &\textcolor{red}{0.80}\\
    \bottomrule
    \end{tabular}}
    \caption{\textbf{Abalation Study on High-quality Sampling Strategy.} The experiments are conducted on GenEval~\citep{ghosh2024geneval}.
    }
    \label{tab:geneval}
\end{table*}

\begin{figure*}[!t]
    \centering
    \includegraphics[width=1.01\linewidth]{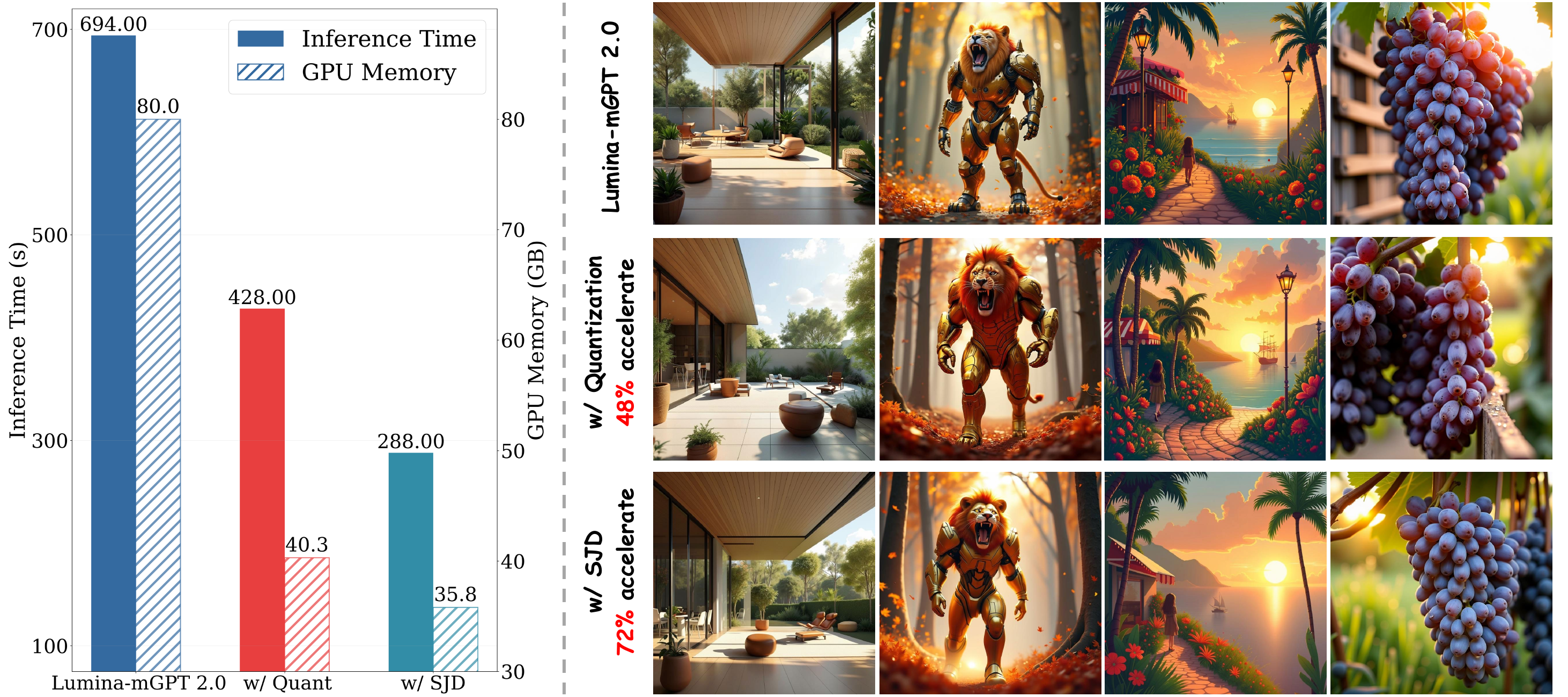}
    \caption{\textbf{Ablation Study on Efficient Sampling Strategy.} The results demonstrate that both SJD and model quantization significantly reduce inference time and GPU memory, while still generating high-quality images.}
    \label{fig:sampling_case}
\end{figure*}

\vspace{0.1cm} \noindent \textbf{Effect of Inference-time Scaling.} 
We integrate inference-time scaling into Lumina-mGPT-2.0 and evaluate its performance in comparison to other large-scale image generation models on the GenEval benchmark, as detailed in Table~\ref{tab:geneval}. By selecting samples from a set of 16 generated images, the inference-scaled model achieves a 11\% increase in overall accuracy compared to naive single-image generation. The improvements are particularly significant in the sub-tasks of ``Two Objects,'' ``Counting'', ``Position,'' and ``Color Attribution.'' These findings highlight that sacrificing inference efficiency can significantly enhance generation quality and accuracy, even when model capacity is constrained.

\vspace{0.1cm} \noindent \textbf{Effect of Acceleration Sampling Strategy.}
In Figure~\ref{fig:sampling_case}, we evaluate the sampling efficiency of Lumina-mGPT 2.0 by integrating model quantization and the speculative jacobi decoding (SJD) strategy. Experimental results indicate that model quantization reduces sampling time by 48\% and GPU memory consumption by 47\% while preserving visual fidelity. Building on this, SJD further enhances efficiency, achieving a 72\% reduction in sampling time through its parallel decoding mechanism. These sampling strategies effectively address the slow sampling speed of Lumina-mGPT 2.0, a common challenge for autoregressive generation models, thereby making it more user-friendly in practical applications.

\section{Conclusion}
\label{sec:conclusion}
This paper introduces Lumina-mGPT 2.0, a standalone, decoder-only autoregressive model for image generation. Lumina-mGPT 2.0 is trained entirely from scratch without incorporating any pre-trained model weights. In the text-to-image generation task, it achieves performance comparable to state-of-the-art models on standard benchmarks, while exhibiting superior visual quality in the synthesized images. Moreover, Lumina-mGPT 2.0 natively supports multiple downstream tasks, enhancing its flexibility and usability for the broader research community.

\vspace{0.1cm} \noindent \textbf{Limitation.} Despite efforts to optimize inference, Lumina-mGPT 2.0 experiences sampling times of several minutes, a challenge shared by all AR-based generation models, which may result in a less-than-ideal user experience. Currently, Lumina-mGPT 2.0 relies on an external LLM for its thinking processes. Future improvements aim to empower the Lumina-mGPT 2.0 to carry out autonomous thinking. Additionally, the current focus of Lumina-mGPT 2.0 is on multimodal generation, upcoming updates are planned to expand its capabilities to include multimodal understanding.

\section*{Acknowledgements}
We thank Yao Teng and Yefei He for their help in accelerating the sampling. We also thank Ziyu Guo’s valuable suggestions for the ``thinking before generation''.

\clearpage
\bibliography{arxiv}

\end{document}